# An iterative algorithm for computed tomography image reconstruction from limited-angle projections

Yuli Sun, Jinxu Tao *, Conggui Liu

*Department of Electronic Engineering and Information Science, University of Science and Technology of China, Hefei 230027, People's Republic of China*

***Abstract:*** In application of tomography imaging, limited-angle problem is a quite practical and important issue. In this paper, an iterative reprojection-reconstruction (IRR) algorithm using a modified Papoulis-Gerchberg (PG) iterative scheme is developed for reconstruction from limited-angle projections which contain noise. The proposed algorithm has two iterative update processes, one is the extrapolation of unknown data, and the other is the modification of the known noisy observation data. And the algorithm introduces scaling factors to control the two processes, respectively. The convergence of the algorithm is guaranteed, and the method of choosing the scaling factors is given with energy constraints. The simulation result demonstrates our conclusions and indicates that the algorithm proposed in this paper can obviously improve the reconstruction quality.

***Keywords:*** Computed tomography, Limited-angle reconstruction, Papoulis-Gerchberg algorithm

## 1. Introduction

Image reconstruction from insufficient data is an important issue in computed tomography (CT). Insufficient data problems occur quite frequently because of practical constraints due to the imaging hardware, scanning geometry, or ionizing radiation exposure [1]. The insufficient data problem may arise from various forms, but in this work we only consider the limited-angle projection problem. From the standpoint of Fourier analysis, the limited-angle problem manifests itself as missing regions in the Fourier space of an image object [2]. Reconstruction by using standard analytic algorithms such as filtered back-projection (FBP) algorithm will lead to distortion and artifact.

---

* Corresponding Author: Tel: +86-551-63601329; Email: tjingx@ustc.edu.cn.

There have been a number of iterative algorithms for overcoming this ill-posed problem. One type of them is based on the compressive sensing (CS) theory proposed by Candes et al. [3]. Main of these algorithms iteratively minimizes the total variation (TV) of the estimated image subject with the constraint that the reconstructed image matches the measured object data. Lots of examples demonstrate that this constrained TV minimization algorithm is effective for limited-angle image reconstruction of objects which have sparse gradients as in [1,4-6]. If the objects have other sparsity, this can also be used for reconstruction in [7,8]. Another type of the iterative algorithms is the iterative reprojection-reconstruction (IRR) algorithm based on the Papoulis-Gerchberg (PG) iteration in [9-11]. The PG algorithm proposed by Gerchberg [12] and Papoulis [13] has been effectively used in various band-limited extrapolation problems such as equivalent currents reconstruction in [14], image super-resolution and inpainting in [15], sparse signal recovery in [16] and limited-angle image reconstruction in [9,10].

It has been proved that the limited-angle image reconstruction can be formulated as an extrapolative problem of band-limited functions in [9]. However, noise is a major problem in the PG algorithm. And the iteration is diverging when the given segment is contaminated with nonband-limited noise [17]. To deal with this problem, Papoulis [13] has suggested the early termination of the iteration; Zhang et al. [18,19] have introduced a scaling factor to make the iteration operator norm less than 1 to ensure the convergence of the algorithm. However, these methods do not fully take into account the effect of noise, and the noisy observation data remains unchanged.

In this article, we present an IRR algorithm using a modified PG iterative scheme to consider the influence of noise. It has two iterative update processes, one is the extrapolation of unknown data, and the other is the modification of the known noisy observation data. And the algorithm introduces three scaling factors to control the two processes, respectively. The proposed algorithm is stable, and we also give the method which is different from [20] to choose the scaling factors with energy constrains.

## 2. Limited-angle problem and Papoulis-Gerchberg algorithm

Let $f \in L^2(R^2)$, the support of $f$ $\text{supp}(f) \subset B = \{x \in R^2 | \|x\| \leq a\}$, which means that the image is in a limited extend. This is in accordance with the CT actual: first, the object to be detected does not fill the whole scanning field; second, due to that the reconstructed image is the external square of the object and the irregular nature of the object, there are some pixels whose value are always zero. We denote this as : $f(x) = f(x) P_B(x)$; where $P_B(x) = \begin{cases} 1 & x \in B \\ 0 & else \end{cases}$.

The imaging model in the X-ray CT, magnetic resonance imaging (MRI) can be described as:

$$\mathcal{R}f(w,p) = \int_{w \cdot x = p} f(x) ds, \qquad (1)$$

where $ds$ is the line element of the line $w \cdot x = p$, $\mathcal{R}f$ is the Radon transform of $f$, is the integral of the image along the straight line $w \cdot x = p$.

By using the Fourier slice theorem, we can obtain $\mathcal{F}_1 \mathcal{R} f = \mathcal{F}_2 f$. If the Radon transform $\mathcal{R}f(w,p)$ is given for all $w$ and $p$, then we can get

$$F(\xi) = \mathcal{F}_2 f = \int_{R^2} f(x) e^{-j2\pi \xi \cdot x} dx, \quad \xi \in \Omega = \{\xi = (p\cos\theta, p\sin\theta) | p \in R, 0 \leq \theta \leq \pi\}.$$

The image $f$ can be reconstructed by $f = \mathcal{F}_2^{-1} \mathcal{F}_1 \mathcal{R} f$ or the filtered back-projection (FBP) algorithm. Here $\mathcal{F}_n$ denote the $n$-dimensional Fourier transform, and $\mathcal{F}_n^{-1}$ denote the $n$-dimensional inverse Fourier transform. However, in the limited-angle problem we can only obtain the partial

$F(\xi), \xi \in \Omega_0 = \{\xi = (p\cos\theta, p\sin\theta) | p \in R, 0 \leq \theta \leq \theta_0 < \pi\}$, and because the $F(\xi)$ is band-limited, we can use the PG algorithm to extrapolate the $F(\xi), \xi \in \Omega/\Omega_0$.

Let $H_0(\xi) = P_{\Omega_0}(\xi)F(\xi)$, where $P_{\Omega_0}(\xi)$ is similar as $P_B(x)$, the PG algorithm for this problem is:

$$\begin{aligned} h_n &= \mathcal{F}_2^{-1} H_n \\ f_n &= h_{n-1} P_B \\ F_n &= \mathcal{F}_2 f_n \\ H_n &= H_0 + (1 - P_{\Omega_0}) F_n \end{aligned} \quad (2)$$

We can get that

$$F_n = H_{n-1} * \mathcal{F}_2\{P_B\}. \quad (3)$$

The PG algorithm iteratively extrapolate the unobserved $F(\xi), \xi \in \Omega/\Omega_0$ by using the observed $F(\xi), \xi \in \Omega_0$ and the band-limited property of $F(\xi)$. Convergence of the algorithm is guaranteed and is shown in [13] under the condition without noise.

From (2) (3), we can get that:

$$f_n = P_B \mathcal{F}_2^{-1} H_0 + P_B \mathcal{F}_2^{-1}\left[(1 - P_{\Omega_0})\mathcal{F}_2 P_B f_{n-1}\right] = f_1 + T_1 f_{n-1}, \quad (4)$$

where $T_1 f = P_B \mathcal{F}_2^{-1}\left[(1 - P_{\Omega_0})\mathcal{F}_2 P_B f\right]$, and with the initial choice $f_1 = P_B \mathcal{F}_2^{-1} H_0$.

In order to ensure the convergence of PG algorithm with noisy projection data, Zhang [18,19] and Qu [9] have proposed a improved PG (IPG) algorithm by introducing a scaling factor $0 < \eta < 1$. Their algorithm is:

$$f_n = f_1 + \eta T_1 f_{n-1}. \quad (5)$$

They have proved the convergence of the IPG algorithm.

## 3. The modified PG algorithm and its convergence

However, when the projection data contain noise, which means that $H_0(\xi) = P_{\Omega_0}(\xi)(F(\xi) + N(\xi))$, where $N(\xi)$ is the noise spectrum. One can find that the influence of noise always exists in the reconstruction. The reason is that $f_1$ remains unchanged in each iteration.

Here we propose a modified PG algorithm which has another iterative update process, it gradually reduce the influence of noise by introduce another two scaling factors $\beta$ and $\gamma$. This algorithm modifies the spectrum of the $\Omega_0$ region in each iteration as:

$$H_n = \beta H_0 + \gamma P_{\Omega_0} F_n + \eta \left(1 - P_{\Omega_0}\right) F_n, \tag{6}$$

where $0 < \beta, \gamma, \eta < 1$

Substituted into (2), we can obtain

$$f_n = \beta P_B \mathcal{F}_2^{-1} H_0 + \gamma P_B \mathcal{F}_2^{-1} P_{\Omega_0} \mathcal{F}_2 P_B f_{n-1} + \eta P_B \mathcal{F}_2^{-1} \left[\left(1 - P_{\Omega_0}\right) \mathcal{F}_2 P_B f_{n-1}\right]. \tag{7}$$

Let $T_1 f = P_B \mathcal{F}_2^{-1} \left[\left(1 - P_{\Omega_0}\right) \mathcal{F}_2 P_B f\right]$, $T_2 f = P_B \mathcal{F}_2^{-1} P_{\Omega_0} \mathcal{F}_2 P_B f$ and $Tf = \eta T_1 f + \gamma T_2 f$, (7) can be written as

$$f_n = \beta f_1 + T f_{n-1}. \tag{8}$$

Next, we will analyze the convergence performance of the proposed algorithm from the perspective of the operator norm.

**Theorem 1.** When $0 < \gamma, \eta < 1$, we can derive that $\|T\| < 1$.

Proof:

$Tf$ can be expressed as three different forms:

First:

$$Tf = \eta T_1 f + \gamma T_2 f,$$

$$\|T\| = \|\eta T_1 + \gamma T_2\| \leq \eta + \gamma. \tag{9}$$

Second:

$$Tf = \eta f + (\gamma - \eta) T_2 f,$$

$$\|T\| \leq \eta + |\gamma - \eta|. \tag{10}$$

Third:

$$Tf = \gamma f + (\eta - \gamma) T_1 f,$$

$$\|T\| \leq \gamma + |\eta - \gamma|. \tag{11}$$

From the above (9) (10) (11), we have $\|T\| < 1$ under the condition of $0 < \gamma, \eta < 1$.

To analyze the final convergence result, as same as [9], we introduce the eigenfunctions $\phi_k(\xi)$ of the equation $\mathcal{F}_2\left[P_B \mathcal{F}_2^{-1}\left(P_{\Omega_0}\phi(\xi)\right)\right] = \lambda \phi(\xi)$ and the corresponding eigenvalues $\lambda_k$ which satisfy

$$1 > \lambda_0 > \lambda_1 > \cdots > \lambda_k > 0; \lambda_k \to 0 \quad as \quad k \to 0,$$

$$\int_{R^2} \phi_i(\xi)\overline{\phi_j(\xi)}d\xi = \delta_{ij}, \quad \int_{\Omega_0} \phi_i(\xi)\overline{\phi_j(\xi)}d\xi = \lambda_i \delta_{ij}. \tag{12}$$

$H_0(\xi)$ can be expanded as a series of $\phi_k(\xi)$:

$$H_0(\xi) = \sum_{k=0}^{\infty} b_k \phi_k(\xi),$$

$$b_k = \frac{1}{\lambda_k} \int_{\Omega_0} H_0(\xi)\overline{\phi_k(\xi)}d\xi. \tag{13}$$

The energy is

$$\|H_0(\xi)\|_{\Omega_0}^2 = \int_{\Omega_0} H_0(\xi)\overline{H_0(\xi)}d\xi = \sum_{k=0}^{\infty} \lambda_k b_k^2. \tag{14}$$

Because $F_n(\xi)$ is band-limited, it can be expanded as

$$F_n(\xi) = \sum_{k=0}^{\infty} a_{n,k} \phi_k(\xi),$$

$$a_{n,k} = \int_{R^2} F_n(\xi)\overline{\phi_k(\xi)}d\xi. \tag{15}$$

**Theorem 2.** The final iteration result is

$$F_*(\xi) = \lim_{n \to \infty} \sum_{k=0}^{\infty} a_{n,k} \phi_k(\xi) = \sum_{k=0}^{\infty} \frac{\lambda_k}{\mu + v\lambda_k} b_k \phi_k(\xi), \tag{16}$$

where $\mu = \dfrac{1-\eta}{\beta}$, $v = \dfrac{\eta - \gamma}{\beta}$.

Proof:

Substituting (3) (6) (13) into (15), we have

$$a_{1,k} = \beta \lambda_k b_k,$$

$$a_{n+1,k} = \beta \lambda_k b_k + \left(\eta + (\gamma - \eta)\lambda_k\right)a_{n,k}. \tag{17}$$

Then we can obtain

$$a_{n,k} = \frac{1 - \left(\eta + (\gamma - \eta)\lambda_k\right)^n}{1 - \left(\eta + (\gamma - \eta)\lambda_k\right)} \beta \lambda_k b_k. \tag{18}$$

It is easy to see that

$$0 < \eta + (\gamma - \eta)\lambda_k < 1. \tag{19}$$

Then we obtain the final iteration result

$$F_*(\xi) = \lim_{n \to \infty} \sum_{k=0}^{\infty} a_{n,k} \phi_k(\xi) = \sum_{k=0}^{\infty} \frac{\beta \lambda_k}{1 - \left(\eta + (\gamma - \eta)\lambda_k\right)} b_k \phi_k(\xi). \tag{20}$$

## 4. Parameters selection with energy constrains

We can ensure the convergence of the proposed algorithm based on the above discussion. However, how to choice the parameters is a difficult problem. An intuitive idea is that we should determine the parameters according to the noise energy. For example, when the noise in $H_0(\xi)$ is slight, obviously the scaling factor $\beta$ should be close to 1 and $\gamma$ should be close to 0.

In this section, we assume that energy bounds on $f(x)$ or $n(x)$ are known. Based upon the additional information, we derive how to choose the scaling factors. To avoid lengthy formulas we adopt the usual norm notation, denoting hereafter

$$\|F(\xi)\|^2 = \int_{R^2} F(\xi)\overline{F(\xi)}d\xi, \tag{21}$$

$$\|F(\xi)\|^2_{\Omega_0} = \int_{\Omega_0} F(\xi)\overline{F(\xi)}d\xi. \tag{22}$$

Given that the energy lower bound and upper bound of $f(x)$:

$$e_L^2 \leq \|F(\xi)\|^2 \leq e_U^2. \tag{23}$$

And the energy lower bound and upper bound of noise in $\Omega_0$:

$$\varepsilon_L^2 \leq \|N(\xi)\|_{\Omega_0}^2 \leq \varepsilon_U^2. \tag{24}$$

**Theorem 3.** We can obtain the following equation:

$$\|H_0(\xi)\|_{\Omega_0}^2 = (2\mu)\|F_*(\xi)\|^2 + (2\nu - 1)\|F_*(\xi)\|_{\Omega_0}^2 + \|H_0(\xi) - F_*(\xi)\|_{\Omega_0}^2. \tag{25}$$

Proof:

Using (13), (14), (16), we have

$$\|H_0(\xi)\|_{\Omega_0}^2 = \sum_{k=0}^{\infty} \lambda_k b_k^2 = \sum_{k=0}^{\infty} \frac{(2\mu)\lambda_k^2 b_k^2}{(\mu + \nu\lambda_k)^2} + \sum_{k=0}^{\infty} \frac{(2\nu - 1)\lambda_k^3 b_k^2}{(\mu + \nu\lambda_k)^2} + \sum_{k=0}^{\infty} \left(1 - \frac{\lambda_k}{\mu + \nu\lambda_k}\right)^2 \lambda_k b_k^2 \tag{26}$$

$$= (2\mu)\|F_*(\xi)\|^2 + (2\nu - 1)\|F_*(\xi)\|_{\Omega_0}^2 + \|H_0(\xi) - F_*(\xi)\|_{\Omega_0}^2$$

Theorem 3 provides us with a method to select the parameters based on the above energy bounds.

A) When $\nu \geq 1/2$, by using (25), we can obtain:

$$\frac{\|H_0(\xi)\|_{\Omega_0}^2 - \|H_0(\xi) - F_*(\xi)\|_{\Omega_0}^2}{\|F_*(\xi)\|^2} \leq 2\mu + 2\nu - 1 \leq \frac{\|H_0(\xi)\|_{\Omega_0}^2 - \|H_0(\xi) - F_*(\xi)\|_{\Omega_0}^2}{\|F_*(\xi)\|_{\Omega_0}^2}. \tag{27}$$

Using (23) and (24), the parameters $\mu$ and $\nu$ should satisfy:

$$2\mu + 2\nu - 1 \geq \frac{\|H_0(\xi)\|_{\Omega_0}^2 - \varepsilon_U^2}{e_U^2}, \tag{28}$$

B) When $\nu < 1/2$, similarly as (27), we have:

$$2\mu + 2\nu - 1 \leq \frac{\|H_0(\xi)\|_{\Omega_0}^2 - \|H_0(\xi) - F_*(\xi)\|_{\Omega_0}^2}{\|F_*(\xi)\|^2} \leq \frac{\|H_0(\xi)\|_{\Omega_0}^2 - \|H_0(\xi) - F_*(\xi)\|_{\Omega_0}^2}{\|F_*(\xi)\|_{\Omega_0}^2}. \tag{29}$$

Using (23), (24), the parameters $\mu$ and $\nu$ should satisfy:

$$2\mu + 2\nu - 1 \leq \frac{\|H_0(\xi)\|_{\Omega_0}^2 - \varepsilon_L^2}{e_L^2}. \tag{30}$$

**Theorem 4.** When $\mu \geq \nu$

$$\|F_*(\xi)\|^2 < \frac{\|H_0(\xi)\|_{\Omega_0}^2}{(\mu + \nu)^2}. \tag{31}$$

When $\mu < \nu$

$$\|F_*(\xi)\|^2 < \frac{1}{4\mu\nu}\|H_0(\xi)\|^2_{\Omega_0}. \tag{32}$$

Proof:

From (16) we have

$$\|F_*(\xi)\|^2 = \sum_{k=0}^{\infty} \frac{\lambda_k}{(\mu+\nu\lambda_k)^2} \lambda_k b_k^2. \tag{33}$$

We denote

$$J(\lambda_k,\mu,\nu) = \frac{\lambda_k}{(\mu+\nu\lambda_k)^2}. \tag{34}$$

And we have

$$\frac{\partial J(\lambda_k,\mu,\nu)}{\partial \lambda_k} = \frac{\mu-\nu\lambda_k}{(\mu+\nu\lambda_k)^3}. \tag{35}$$

Substituting $0 < \lambda_k < 1$, we can obtain:

When $\mu \geq \nu$,

$$J(\lambda_k,\mu,\nu) < J(\lambda_k = 1,\mu,\nu) = \frac{1}{(\mu+\nu)^2}. \tag{36}$$

When $\mu < \nu$,

$$J(\lambda_k,\mu,\nu) < J\left(\lambda_k = \mu/\nu, \mu, \nu\right) = \frac{1}{4\mu\nu}. \tag{37}$$

Based on (23) and Theorem 4, we have another method to select the parameters:

If $\dfrac{\|H_0(\xi)\|^2_{\Omega_0}}{(\mu+\nu)^2} \leq e_L^2$; then the $\mu$ and $\nu$ we selected should satisfy: $\mu < \nu$ and $\mu\nu \leq \dfrac{\|H_0(\xi)\|^2_{\Omega_0}}{4e_L^2}$. Otherwise the $\mu$ and $\nu$ only need to satisfy

$$(\mu+\nu)^2 < \frac{\|H_0(\xi)\|^2_{\Omega_0}}{e_L^2}.$$

## 5. Simulation result

In this section, simulation is performed to demonstrate the proposed conclusions and evaluate the performance of the modified PG algorithm. Here we consider the realistic case of distorted observation. Particularly, samples are taken from noisy versions of the $256 \times 256$ Shepp-Logan phantom test image, affected by additive white Gaussian noise (AWGN).

Figure 1 shows the reconstructions for limited angle scans covering $2\pi/3$ by using the direct DFT, PG algorithm, IPG algorithm and the proposed algorithm, respectively. And we use the gridding method as Zhang et al. [21]. From Figure 1(d), one can find that the PG algorithm is diverging in this condition. Comparing Figure 1(f) with other reconstructed images, one can find that our proposed algorithm not only can ensure the convergence of the iteration but also can improve the quality of reconstruction.

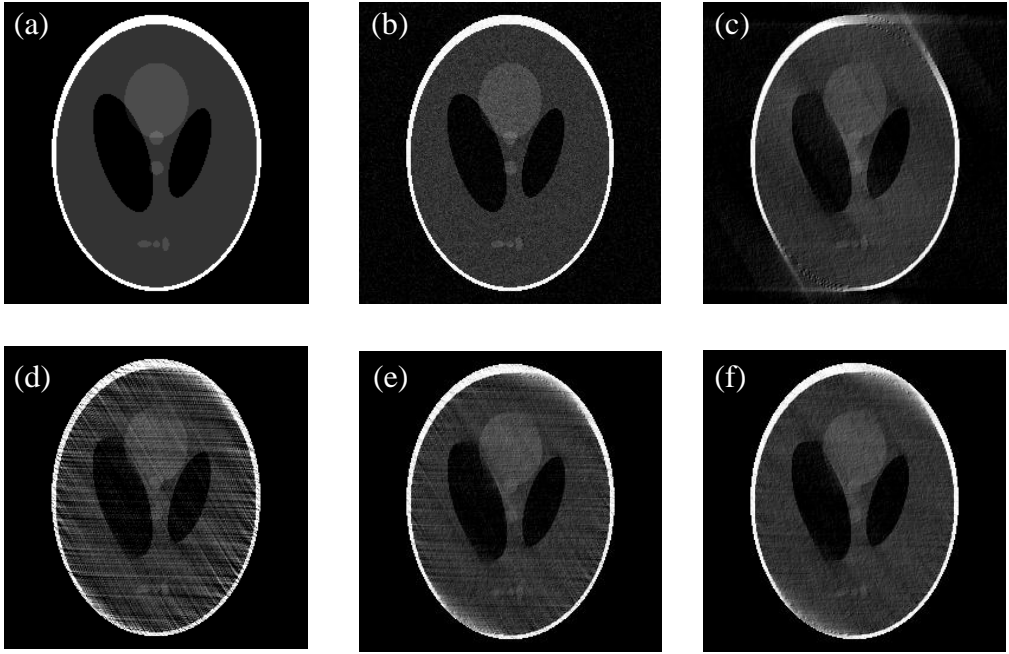

**Figure 1**. Reconstructions from noisy Shepp-Logan phantom affected by AWGN. (a) Original Shepp-Logan phantom. (b) Noisy phantom, SNR is 15dB. From (c) to (f), corresponding to the reconstructed images based on direct DFT, PG algorithm, IPG algorithm and the proposed algorithm, respectively.

Here we also use three objective evaluation parameters to evaluate the reconstructed image: the normalized mean square distance ($d$), the normalized absolute average distance ($r$) [22], and the peak signal to noise ratio ($PSNR$).

$$d = \left( \frac{\sum_{i=1}^{N}\sum_{j=1}^{N}(t_{i,j}-f_{i,j})^2}{\sum_{i=1}^{N}\sum_{j=1}^{N}(t_{i,j}-\bar{t})^2} \right)^{1/2}, \quad r = \frac{\sum_{i=1}^{N}\sum_{j=1}^{N}|t_{i,j}-f_{i,j}|}{\sum_{i=1}^{N}\sum_{j=1}^{N}|t_{i,j}|}.$$

Where $t$ is the test standard image, $\bar{t}$ is the average value of $t$, $f$ is the reconstructed image. We can see that $d$ is sensitive to big errors, while $r$ is sensitive to whole errors.

From table 1 it can be drawn that the performance of our proposed algorithm is better than other algorithms. One can also find that when the SNR increases, the performance improvement decreases. This is due to the fact that the proportion of noise in the observed $\Omega_0$ region becomes smaller, then the scaling factor $\beta$ should gradually close to 1 as we have mentioned previously. This can also be explained by using (25), when the SNR increases, the sum of $u$ and $v$ is close to 1, which means that the proposed algorithm gradually degenerated into IPG algorithm by using (16).

**Table 1**. Performance comparison of different algorithms of different noise levels.

| SNR | Parameters | PG | IPG | Proposed |
|---|---|---|---|---|
| 5dB | d | 1.3988 | 0.5772 | 0.4457 |
| | r | 1.1943 | 0.5162 | 0.4146 |
| | PSNR | 10.4637 | 18.1527 | 20.3988 |
| 10dB | d | 0.7905 | 0.3612 | 0.3045 |
| | r | 0.6768 | 0.3135 | 0.2643 |
| | PSNR | 15.4210 | 22.2246 | 23.7072 |
| 15dB | d | 0.4603 | 0.2607 | 0.2405 |
| | r | 0.3933 | 0.2075 | 0.1804 |
| | PSNR | 20.1177 | 25.0555 | 25.7063 |
| 20dB | d | 0.2949 | 0.2292 | 0.2169 |
| | r | 0.2424 | 0.1626 | 0.1521 |
| | PSNR | 23.9862 | 26.5602 | 26.8519 |

## 6. Conclusion

In this paper, we have developed and applied an algorithm based on a modified PG iterative scheme for the practical reconstruction from limited-angle projections which contain noise. We have proved the convergent property of the proposed algorithm and provided the method of choosing parameters with energy constraints. The simulation results show that the proposed algorithm can obtain a robust and better reconstructed image. Future studies include extending our analysis to 3D imaging, combining it with other reconstruction algorithms.